\documentclass[journal, comsoc]{IEEEtran}

\ifCLASSINFOpdf \else \fi
    
\usepackage{graphicx} 
\usepackage{dblfloatfix} 
\usepackage{graphicx} 
\usepackage{array}    
\usepackage{booktabs} 
\usepackage{amsmath,amssymb}
\usepackage{cite}
\usepackage{amsfonts}
\usepackage{amsmath}
\usepackage{algorithmicx}
\usepackage{graphicx}
\usepackage{subcaption}
\usepackage{algpseudocode}
\usepackage{xcolor}
\usepackage{tikz}
\usepackage{pgfplots}
\usepgfplotslibrary{statistics}

\usepackage{balance}
\usepackage{algorithm}
\usepackage{mathtools}
\usepackage{comment}
\newcommand{\bm}[1]{\boldsymbol{#1}}

\begin{document}

\title{A novel RF-enabled Non-Destructive Inspection Method through Machine Learning and Programmable Wireless Environments}

\author{Stavros Tsimpoukis,
Dimitrios Tyrovolas,~\IEEEmembership{Member,~IEEE,}
Sotiris Ioannidis, Maria Kafesaki,\\ Ian F. Akyildiz,~\IEEEmembership{Life Fellow,~IEEE,}
George K. Karagiannidis,~\IEEEmembership{Fellow,~IEEE,}
Christos K. Liaskos
\thanks{S. Tsimpoukis is with the University of Ioannina, Ioannina, Greece (s.tsimpoukis@uoi.gr).}
\thanks{D. Tyrovolas is with the 
Department of Electrical and Computer Engineering, Aristotle University of Thessaloniki, 54124 Thessaloniki, Greece, and with Dienekes SI IKE, 71414 Heraklion, Greece (tyrovolas@auth.gr).}  
\thanks{S. Ioannidis is with the Department of Electrical and Computer Engineering, Technical University of Chania, Chania, Greece, and with Dienekes SI IKE, 71414 Heraklion, Greece (sotiris@ece.tuc.gr).}
\thanks{M. Kafesaki is with the Foundation for Research and Technology Hellas (FORTH), Greece (kafesaki@iesl.forth.gr).}
\thanks{I. F. Akyildiz is with Truva Inc., Alpharetta, GA 30022, USA (ian@truvainc.com).}
\thanks{G. K. Karagiannidis is with the Department of Electrical and Computer Engineering, Aristotle University of Thessaloniki, 54124 Thessaloniki, Greece (geokarag@auth.gr).}
\thanks{C. K. Liaskos is with the Computer Science Engineering Department, University of Ioannina, Ioannina, and Foundation for Research and Technology Hellas (FORTH), Greece (cliaskos@ics.forth.gr).} 
}

\IEEEtitleabstractindextext{%
\begin{abstract}

Contemporary industrial Non-Destructive Inspection (NDI) methods require sensing capabilities that operate in occluded, hazardous, or access restricted environments. Yet, the current visual inspection based on optical cameras offers limited quality of service to that respect. In that sense, novel methods for workpiece inspection, suitable, for smart manufacturing are needed. Programmable Wireless Environments (PWE) could help towards that direction, by redefining the wireless Radio Frequency (RF) wave propagation as a controllable inspector entity. In this work, we propose a novel approach to Non-Destructive Inspection, leveraging an RF sensing pipeline based on RF wavefront encoding for retrieving workpiece-image entries from a designated database. This approach combines PWE-enabled RF wave manipulation with machine learning (ML) tools trained to produce visual outputs for quality inspection. Specifically, we establish correlation relationships between RF wavefronts and target industrial assets, hence yielding a dataset which links wavefronts to their corresponding images in a structured manner. Subsequently, a Generative Adversarial Network (GAN) derives visual representations closely matching the database entries. Our results indicate that the proposed method achieves an SSIM $99.5\%$ matching score in visual outputs, paving the way for next-generation quality control workflows in industry.

\end{abstract}

}

\maketitle

\IEEEdisplaynontitleabstractindextext

 \section{Introduction}

The contemporary manufacturing process is characterized by the emergence of Cyber-Physical Systems (CPS), and Digital Twins (DTs) which envision a highly-interconnected and automated industrial era \cite{lu2020digital, cps}. Vital to realizing this paradigm is the utilization of Non-Destructive Inspection (NDI) technologies, which monitor the manufacturing environment and provide structural visualizations related to the physical assets, without disrupting the production line \cite{jaber2022smart}. Achieving consistent production quality depends on the environment's ability to monitor, in real-time, the manufacturing process and infer accurate decisions about products' quality, placing strict demands on system speed, coordination and reliability. To meet these demands, current approaches rely on a network of interconnected hardware, where cameras and sensors capture continuous inputs and computing units deduce on products' structural integrity. However, optical methods of inspection often fall short in meeting industry-imposed quality standards because of the inherent limitations related to line-of-sight (LoS) requirements, occlusion, and lighting contamination issues. Additionally, due to the transition of the manufacturing sector towards industry 5.0, the focus of automation should be enriched with systems that consider a human-centric rationale \cite{xu2021industry}. This paradigm drives the inspection endeavor not only to ensure a high-fidelity manufacturing process, but also respect workforce privacy, which is an aspect that camera-based monitoring fails to consider. Therefore, it becomes crucial to investigate alternative approaches which can both satisfy an automated NDI system that ensures a reliable visual monitoring, while also respecting constraints related to strict human-centric privacy requirements.

A promising step toward decoupling the environment's need to rely on camera-based  processing approaches involves shifting the inspection aspect to the physical propagation of abundant radio-frequency (RF) waves within the environment. In more detail, since RF waves travel along the environment and interact with it, manipulating their behavior during transmission allows key information to be encoded directly into the physical RF signal, reducing reliance on extensive, multi-component digital processing, thus offering a new sensing modality. However, realizing this concept requires the ability to dynamically shape how RF signals interact with their surrounding environment, turning the propagation medium itself into an active part of the system's inspection mechanism. In this direction, recent advances in engineered materials known as metamaterials have made such control feasible \cite{liaskos2020internet, kaina2014shaping, kossifos2020toward, liu2019intelligent, Mekikis2022}. Specifically, these materials are built from periodic arrangements of sub-wavelength elements, known as meta-atoms, which can be tailored to induce precise and controllable interactions with incident waves. A refined implementation of this concept is found in reconfigurable intelligent surfaces, also known as software-defined metasurfaces (SDMs), which extend this control to real-time operation by enabling electronic configuration of each element to adjust how signals are reflected, refracted or absorbed \cite{Mekikis2022,tsilipakos2020toward}. Therefore, when such metasurfaces are applied to the surfaces of a physical space and configured cooperatively, they can convert the wireless propagation environment into a programmable wireless environment (PWE) \cite{liaskos2018new}, where electromagnetic (EM) wave propagation becomes a fully controllable resource, carrying useful information. As a result, the environment itself can support NDI functions during signal propagation, thus redefining how visual information, related to physical assets' appearance, is acquired and processed.

Given the ability of PWEs to manipulate EM signal propagation in real time, it becomes feasible to shift the inspection intelligence into the physical layer, minimizing dependence on conventional optical sensing hardware. RF waves present a particularly suitable option, as they can propagate through complex environments and interact with physical assets in a way that inherently encodes spatial and structural information. Additionally, RF waves allow for an invasive yet not harmful monitoring of physical assets based on the nature of the EM waves to penetrate materials and encode information for total material-aware volumetric composition, a capability not supported by the current camera inspection methods.  At the same time, PWEs can guide this process by shaping how wavefronts form and propagate, embedding useful transformations during transmission, and enhancing the informativeness of the captured signal. Building on this principle, an automated NDI system can be envisioned where an RF source illuminates the inspection zone, and the target asset within it naturally alters the propagation of the signal through scattering. This enables the extraction of physical information and the generation of high-fidelity DTs of the assets for a-posteriori further monitoring and processing. This naturally enables an industrial extended reality (XR) environment \cite{liaskos2022xr}, which aids remote operators to inspect hazardous or inaccessible zones, without a local physical presence. To this end, this approach suggests a compelling alternative to traditional computer vision pipelines, by guaranteeing the preservation of privacy, and invariance to illumination conditions.

Motivated by the ability of PWEs to control signal propagation and by the capacity of RF wavefronts to carry spatial and structural information, this work proposes a method for embedding DT reconstruction directly into the physical layer using PWE-driven signal shaping and Generative Adversarial Networks (GANs). Specifically, by routing RF signals through a controlled environment, the PWE enables the formation of wavefront patterns that are uniquely associated with each industrial asset in the scene and are deliberately structured to reflect the features needed for their visual reconstruction, establishing a meaningful link between RF signals and the asset's visual representation. Furthermore, when these wavefronts are captured by an array of receivers, the resulting measurements serve as a physically encoded input to a GAN trained to learn the mapping from wavefront structure to high-fidelity inspection images, enabling accurate remote monitoring without conventional tracking or rendering pipelines. Finally, simulation results show that the proposed approach achieves high accuracy in retrieving correct visual representations from a reference asset-database, confirming the potential of PWE-controlled RF encoding as a foundation for novel NDI systems. Thus, the main contribution of this paper can be summarized in the following.
\begin{itemize}
    \item 
    Propose a PWE-driven RF-encoding methodology for DT reconstruction purposes.
    
    \item 
    Accurate visualization of geometric shapes and structural variations.
    
    \item 
    Operational resilience in arbitrary illumination conditions, and a privacy preserving pipeline.
    
\end{itemize}

The remainder of this paper is organized as follows. Section II reviews related work, while Section III presents the system model and outlines the proposed methodology. Section IV introduces the similarity metrics used to associate RF wavefronts with visual representations and details the construction of the corresponding RF signals. Section V describes the routing strategy for generating workpiece-specific RF readings within the programmable environment, while Section VI provides numerical results. Finally, Section VII concludes the paper.

\section{Related Work}

The pursuit of ever increasing, high-quality Non-Destructive Inspection systems has been widely recognized as a central challenge, particularly as the contemporary industrial era strives for rigorous quality assurance objectives. To address this, several works have proposed solutions that aim to improve the automated inspection technologies that make up the NDI pipeline. Specifically, to meet industry guidelines of continuous quality assurance, the majority of studies focus on harnessing optical visual inspection for delivering an automated NDI technique within industrial settings. State-of-the art studies examine the utilization of the modern data-driven paradigm, via machine learning (ML), and thus examine the co-operation between standard camera-based inspection and deep-learning inference. In particular \cite{rovzanec2022towards}, studies the use of ML in order to perform surface anomaly assessment during manufacturing. However, such optical-based methods are inherently limited by their reliance on LoS connection to the industrial assets, while also falling short to ensure industry-imposed quality standards when considering occlusion, or lighting degradation conditions such as smoke or low illumination, a typical scenario to industrial manufacturing. Additionally, in order to tackle these limitations, alternative NDI methods, such as ultrasonic testing (UT), \cite{jodhani2023ultrasonic} have been adopted. In essence, these methods utilize sound waves to inspect the various industrial assets, while also allowing for internal inspection. Nevertheless, this method necessitates a physical contact, or coupling medium for accurate quality deduction, thus making it non-optimal for real-time inspection of moving products along the manufacturing line. Yet, even in cases of non-contact variants these methods face difficulties due to high signal attenuation, and struggle with complex asset geometries. Furthermore, an additional method considers the utilization of X-ray inspection \cite{x_ray}, which provides a high-resolution capability to monitor the manufacturing process. Yet, this method raises significant safety concerns when it does not guarantee strict radiation shielding, hence not offering an optimal human-centric alternative. As a result, there is an emerging need for finding alternative sensing modalities that can overcome these limitations and ensure high-quality and human-friendly asset inspection.

To address the systemic challenges and physical limitations of traditional NDI systems, inspection by harnessing RF signals has emerged as a promising novel alternative. In contrast, to the lighting-related and privacy issues of optical hardware, medium requirements of UT, and safety concerns related to X-ray, RF-signal utilization for remote monitoring offers a private, wireless, and safe alternative to the industry inspection endeavor. Specifically, the family of RF tomography methods in \cite{styla2023image, klosowski2022use, wu2020convolutional, wang2015enhancing} estimate object positions by analyzing how RF signals traverse indoor environments, often using ML to support coarse object localization. Furthermore, radar-based approaches in \cite{deshmukh2022physics, liu2024dynamic, 7916455} aim to capture coarse object geometries from reflected signals, or even physics-assisted neural models \cite{deshmukh2022physics}, produce coarse reconstructions. Moreover, recent advancements in the RF-sensing domain demonstrated the ability for human activity recognition, \cite{koutsonikolas}, and human body pose estimation, \cite{mit_body_posture}, by making use of a synergy workflow between RF-signals and ML. The main advantage of these works is the ability to operate in relative low spectrum frequencies, i.e., WiFi bands at 5GHz, while achieving adequate efficiency in visual representation. Although all of the aforementioned methods highlight the value of RF signals for environmental perception, they are primarily limited to localization or low-resolution imaging and do not support the generation of high-quality visual content, hence making them unsuitable for a precise automated NDI system which can ensure high-quality monitoring. To move beyond passive sensing, additional efforts have focused on actively shaping signal propagation through PWEs. It is important to note the fundamental distinction between the traditional radar-based imaging approach, and the proposed PWE-driven method. The core notion of the radar systems lies in the high-bandwidth pulse transmission and the digital post-processing of the received echo signals. As a result, this approach infers location and geometrical cues simply by the time-of-flight related information. On the contrary, the PWE method proposes the conversion of the environment, itself, into an auxiliary "RF-lens" to highlight the RF features corresponding to the illuminated assets. By programming the scattering inside the PWE, it is possible to enrich the information encapsulated within the RF signals and utilize various degrees-of-freedom, other than the time-of-arrival related power, e.g., polarization etc. This shifts the perspective from a-posteriori RF-echo digital post processing, to a physical-layer wave control, realizing high-quality visual reconstruction without the need for ultra-wideband spectrum. In particular, the authors in \cite{cscn2023} examine how different PWE configurations affect RF-based object classification within a controlled propagation environment. However, this work does not define a structured mechanism for encoding visual information into RF signals, nor does it demonstrate how to reconstruct rich visual content suitable for NDI applications. Thus, to the best of the authors’ knowledge, no complete framework currently exists for generating visual inspection data directly from wireless signals within a PWE, marking a key gap in the literature. 

\section{System Model \& Methodology}

\begin{figure*}[t!]
\centering{} {
 \includegraphics[clip,width=0.95\linewidth]{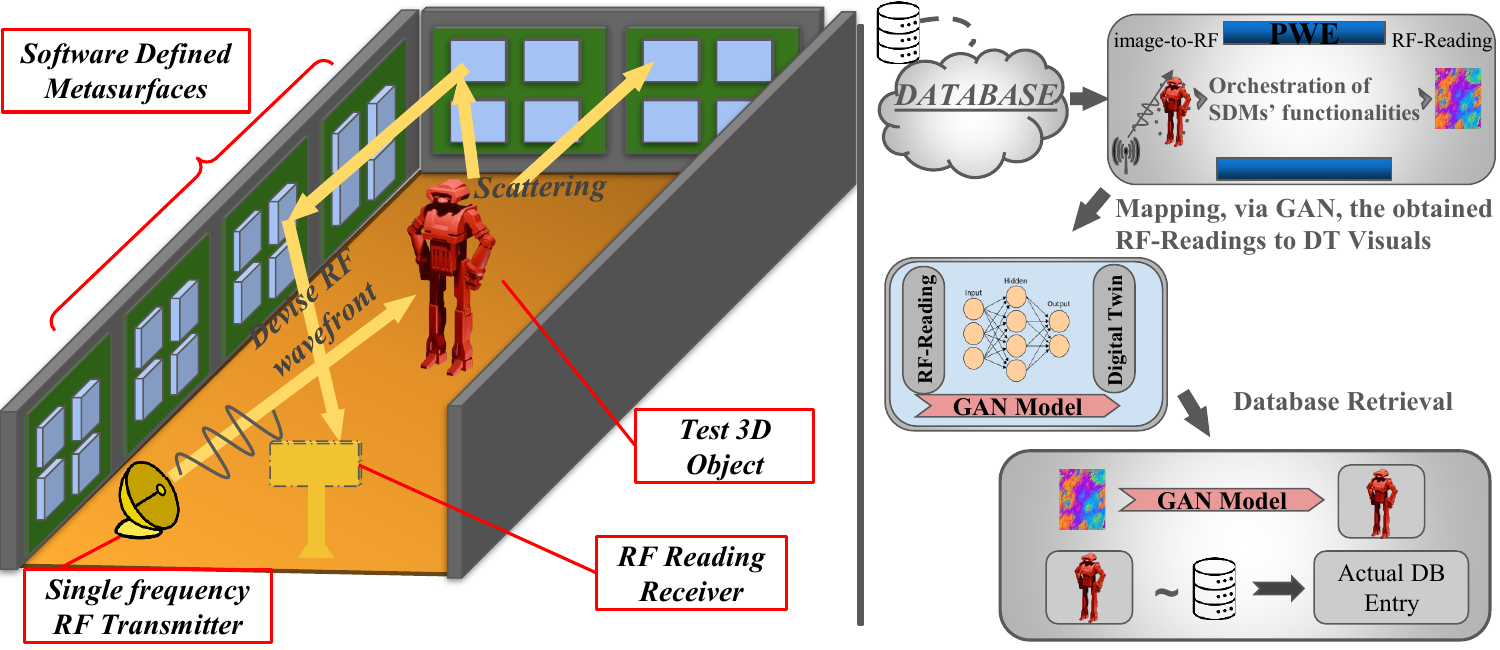}}
\caption{\protect\label{fig:System Model} Overview of the NDI process via RF-Encoding manipulations, within an indoor PWE. While the right visualization illustrates an abstract representation of the overall RF-Wavefronts to Digital Twin visuals workflow.}
\end{figure*}

\subsection{System Model}
We consider a PWE deployed within an enclosed indoor space, creating an inspection chamber, as shown in Fig. \ref{fig:System Model}. Single-frequency transmitters emit RF signals, an industrial asset selected from a predefined database of workpieces is placed at the center of the room, and a receiver equipped with multiple antennas is positioned to collect the scattered wavefronts. The environment is designed to control the propagation of these signals such that the structure of the received wavefronts encodes geometry-dependent characteristics of the object. To enable this behavior, the walls and ceiling are coated with SDMs, which dynamically manipulate the EM response of impinging RF signals. Specifically, each SDM selects its behavior from a predefined codebook, where each entry corresponds to a specific wave manipulation functionality. The codebook adheres to a physics-informed training which captures the underlying EM phenomena, i.e., unit-cell mutual-coupling and action-dependent phase adjustment, for accurately configuring the SDMs' macroscopic functionality \cite{Papadop2024Codebook}. In this setting, two core functionalities are considered: i) beam steering, which directs incoming signals toward a designated SDM or receiver antenna under line-of-sight (LoS) conditions, and ii) absorption, which suppresses undesired reflections by absorbing the impinging waves. Thus, by coordinating these functionalities across the environment, the PWE shapes the scattered signals produced by the object into directed wavefronts that reach the receiver with spatial patterns that reflect the structure of the object, forming a compact and informative signal representation suitable for subsequent visual assessment.

\subsection{Methodology}\label{sec:32}
The proposed methodology establishes a signal-to-image pipeline that generates visual surface reconstructions of industrial assets from structured RF wavefronts shaped by the PWE. This mapping is realized through a supervised learning framework in which a generative model is trained to associate received signal patterns with corresponding target images. These target images are grayscale renderings of the object, captured under fixed lighting and consistent perspective. In this context, each signal pattern, referred to as an RF-Reading, captures spatial and propagation-related features that reflect how EM waves interact with the object and traverse the environment. Hence, the dataset required for the supervised training is constructed by pairing RF-Readings with target images that represent the same object observed under the same scene conditions. As a result, the generative model, implemented as a GAN, learns to synthesize visual content directly from the structured characteristics of the received signals.

To support this learning process, a supervised dataset is constructed, where each training sample consists of an RF-Reading vector $\boldsymbol{w}$ paired with the corresponding object image. Specifically, $\boldsymbol{w}$ encapsulates the directional and propagation-related measurements collected at the receiver and is expressed as
\begin{equation}\label{eq:w}
    \boldsymbol{w} = [\phi_1, \theta_1, l_1, \psi_1, \dots, \phi_m, \theta_m, l_m, \psi_m],
\end{equation}
where $m$ denotes the number of antennas in the receiver array. Each group of four consecutive elements in $\boldsymbol{w}$ corresponds to a single antenna and includes the azimuth angle of arrival $\phi_i$, the elevation angle of arrival $\theta_i$, the path loss $l_i$, and the impinging phase $\psi_i$ observed at the $i$-th receiving element. Among these, the direction-of-arrival (DoA) is defined for each antenna as
\begin{equation}\label{eq:DoA}
    \mathrm{DoA}_i = \{\phi_i, \theta_i\},
\end{equation}
which describes the incoming signal’s orientation in space relative to the corresponding antenna. As a result, these directional features provide a spatial signature of the wavefront and serve as a key component in learning the mapping from RF measurements to visual reconstructions.

To enhance the accuracy of the RF-to-image mapping, the PWE is configured not only to guide RF signals toward the receiver, but also to impose deliberate structure on the wavefronts. Specifically, the system is designed so that visually similar objects produce RF-Readings with similar directional patterns. This is accomplished by orchestrating the SDM functionalities to control the signal’s propagation paths such that the DoA features reflect object-level similarity relationships found in the image domain. As a result, the system encodes object-discriminative spatial patterns into the RF domain, allowing the model to better learn consistent associations between signal and image, as also supported by established principles in pattern recognition \cite{bishop2006pattern}. However, even if this structured design promotes a clearer relationship between RF-Readings and visual outputs, the discrete nature of SDM codebooks prevents reliable deterministic mappings. To address this, the generative model is implemented as a GAN that learns to synthesize plausible object images based on imperfect but feature-rich signal inputs. In more detail, through supervised training, the GAN models the latent structure of the directional and propagation features in $\boldsymbol{w}$ and learns to infer visual representations that preserve shape-related information despite noise and distortion.

After training, the system transitions to an evaluation phase, where a previously unseen object is introduced into the scene under identical propagation conditions. At this stage, the PWE applies an SDM configuration synthesized by aggregating multiple precomputed responses corresponding to distinct object orientations, as detailed in \cite{segata2024cooperis}. This unified configuration enables the generation of an RF-Reading that encapsulates directional information without requiring new SDM coordination for each test instance. The obtained signal is subsequently translated into a visual representation through the trained GAN model, which infers visual characteristics directly from the latent structure of the wavefront. Once the candidate image is produced, it is compared against a reference set of stored various asset images using a similarity metric-based retrieval process. The closest match is identified and selected as the final output, effectively linking the received RF-wavefront to a known visual surface representation. Within the broader Industry 4.0 architecture, this module is positioned to operate alongside an upstream object-type classifier that assists in policy selection for SDM behavior \cite{cscn2023}, and can be followed by predictive maintenance systems, feeding real-time visual data to an automated inspection system.

\section{RF-Encoding Process}
To generate RF-Readings that meaningfully reflect the visual characteristics of objects, it is important to consider how the PWE can influence the structure of the received signals. Among the various signal properties that could potentially carry such information, the direction-of-arrival stands out as a practical choice. Unlike other RF parameters that may require complex, multi-dimensional tuning, the directionality of the received wavefronts can be reliably shaped by adjusting the behavior of the environment’s reconfigurable elements. This makes it possible to guide the incoming signals in a way that visually similar objects produce similar directional patterns at the receiver. Such alignment between signal structure and visual content plays a critical role in facilitating the learning task that connects RF-Readings to object images. To enable this alignment in a systematic manner, the first step is to define a reliable way of measuring how similar two images are, so that this information can then be embedded into the structure of the signals themselves. In this direction, this section analyzes how visual similarity can be expressed through a vector-based metric, and how this metric serves as a guide for generating directional signal features that preserve such relationships within the RF domain.

\subsection{Similarity Metric}

To enable consistent and accurate signal-to-image mappings, it is essential to construct RF-Readings in a way that preserves the visual relationships shared among the corresponding object images. In more detail, since the RF-Readings are represented as structured vectors that capture directional propagation characteristics, their design must reflect similarity patterns that are meaningful in both the visual and RF domains, thus ensuring that objects with similar visual appearances produce RF-Readings with similar structural traits. To support this goal, the object images must first be transformed into vector representations that are compatible with the vector nature of the RF-Readings, not only in structural form but also in terms of numerical consistency, since each RF-Reading consists of scalar-valued entries. In more detail, as each image can be inherently represented as a two-dimensional array of grayscale pixel intensities, this transformation is realized by flattening the matrix into a one-dimensional vector, thereby preserving the image content while aligning it structurally and numerically with the directional signal components. Thus, this representation enables a consistent basis for measuring visual similarity in a form directly comparable to the directional encoding in the RF domain, forming the foundation for the construction of the RF-Readings.

Based on this formulation, the selection of an appropriate similarity metric becomes a central design decision, as it directly affects how visual relationships are translated into RF-Readings. In this direction, the Pearson correlation coefficient can be adopted, as it provides a robust means of quantifying the linear alignment between pairs of vectorized images. In particular, given a set of target images, each flattened into a one-dimensional vector $\boldsymbol{g}^{i}$, their pairwise correlations are computed to form a correlation matrix $\boldsymbol{R}$ defined as
\begin{equation}\label{eq:R}
\boldsymbol{R} = 
\begin{pmatrix}
r_{1,1} & r_{1,2} & \cdots & r_{1,n} \\
r_{2,1} & r_{2,2} & \cdots & r_{2,n} \\
\vdots  & \vdots  & \ddots & \vdots  \\
r_{n,1} & r_{n,2} & \cdots & r_{n,n} 
\end{pmatrix},
\end{equation}
where each entry $r_{i,j}$ is given by
\begin{equation}
r_{i,j} = \frac{\sum_{k=1}^{n} (g_{k}^{i}-\overline{g^{i}})(g_{k}^{j}-\overline{g^{j}})}
{\sqrt{\sum_{k=1}^{n} (g_{k}^{i}-\overline{g^{i}})^2 \sum_{k=1}^{n} (g_{k}^{j}-\overline{g^{j}})^2}},
\end{equation}
with $g_k^i$ and $g_k^j$ denoting the $k$-th pixel of the $i$-th and $j$-th images, respectively, and $\overline{g^i}$ the mean pixel value of image $i$. As a result, $\boldsymbol{R}$ can serve as the reference model of visual similarity, thus, obtaining an interpretable basis for constructing RF-Readings whose directional components can be aligned to mirror the visual similarity behavior encoded in $\boldsymbol{R}$.

\subsection{RF-Encoding} 
To construct RF-Readings that reflect the visual similarity structure observed among object images, the directed signals within the PWE must be synthesized to exhibit correlation patterns that mirror those found in the visual domain. This goal can be addressed through the formulation of a matrix $\boldsymbol{D} \in \mathbb{R}^{n \times m}$, which can be expressed as
\begin{equation}
    \boldsymbol{D} = 
\begin{pmatrix}
\mathrm{DoA}_{1,1} & \mathrm{DoA}_{1,2} &\cdots & \mathrm{DoA}_{1,m} \\
\mathrm{DoA}_{2,1} & \mathrm{DoA}_{2,2} & \cdots & \mathrm{DoA}_{2,m} \\
\vdots  & \vdots  & \ddots & \vdots  \\
\mathrm{DoA}_{n,1} & \mathrm{DoA}_{n,2} & \cdots & \mathrm{DoA}_{n,m} 
\end{pmatrix},
\end{equation}
where each element corresponds to a DoA vector at the $m$-th antenna of the receiver for each of the $n$ objects of the data base. To ensure that the inter-object relationships embedded in $\boldsymbol{D}$ align with the visual similarity structure, an empirical correlation matrix $\hat{\boldsymbol{R}}(\boldsymbol{D})$ needs to be computed based on the entries of $\boldsymbol{D}$ to approximate the reference matrix $\boldsymbol{R}$ derived from the vectorized image representations. To this end, this can be achieved by solving an optimization problem that minimizes the Frobenius norm $||\cdot||_{F}$ of the difference between $\hat{\boldsymbol{R}}(\boldsymbol{D})$ and $\boldsymbol{R}$, which provides a scalar measure of total deviation across all matrix entries and promotes a balanced alignment of pairwise correlations. Each element $d_k^i$ of $\boldsymbol{D}$ is constrained to lie within the interval $[0, 1]$, enabling a unified encoding of azimuth and elevation components without the need to separately handle their differing native ranges. This normalization promotes fairness when minimizing across the entries of $\boldsymbol{D}$, ensuring that no directional axis disproportionately influences the learned similarity structure. Subsequently, these normalized values are mapped back to valid DoA angles through a decoding step that re-scales them to their corresponding physical ranges, making the optimized directions suitable for realization through SDM steering configurations. Thus, the resulting optimization problem is expressed as
\begin{equation} \label{eq:opt_problem}
\begin{aligned}
\underset{\boldsymbol{D} \in [0, 1]^{n \times m}}{\text{minimize}} \quad & \left|\left| \hat{\boldsymbol{R}}(\boldsymbol{D}) - \boldsymbol{R} \right|\right|_{F},
\end{aligned}
\end{equation}
where $\hat{\boldsymbol{R}}$ is computed using the Pearson correlation formula:
\begin{equation} \label{eq:pearson}
\hat{r}_{i,j} = \frac{\sum_{k=1}^{m} (d_k^i - \overline{d^i})(d_k^j - \overline{d^j})}{\sqrt{\sum_{k=1}^{m} (d_k^i - \overline{d^i})^2 \sum_{k=1}^{m} (d_k^j - \overline{d^j})^2}},
\end{equation}
with $d_k^{i}$ corresponding to the $k$-th element of matrix $\boldsymbol{D}$ in the $i$-th row, and $\overline{d^i}$ denoting the mean of the $i$-th row. This formulation ensures that the learned signal structure reliably encodes the correlation-based semantics of the visual domain.

To initialize the optimization process with a statistically informed prior, the initial DoA matrix $\boldsymbol{D}_0$ is sampled from a multivariate normal distribution $\mathcal{N}(\boldsymbol{0}, \boldsymbol{R})$, where the covariance structure reflects the target correlation matrix. This initialization generates DoA samples whose interrelationships approximate the visual domain similarity structure, improving convergence efficiency. The optimization proceeds iteratively,
by updating the rows of $\boldsymbol{D}$ based on a strategy that performs serial updates on the rows of the matrix. Each selected DoA vector is refined using the L-BFGS-B algorithm, which handles bound constraints while preserving efficiency for high-dimensional problems \cite{liu1989limited}. This iterative refinement continues for a fixed number of outer iterations, progressively aligning $\hat{\boldsymbol{R}}$ with $\boldsymbol{R}$, as detailed in Algorithm \ref{alg:rf_vector_opt}. Although the described approach is based on numerical optimization, it is worth noting that alternative formulations such as Cholesky decomposition or eigenvalue-based decomposition could also serve as approximate solutions for generating a correlation-consistent $\boldsymbol{D}$ matrix, although with reduced flexibility in handling element-wise constraints.

Once the optimization process produces a DoA matrix $\boldsymbol{D} \in [0, 1]^{n \times m}$ whose correlation structure aligns with the visual similarity captured in the image domain, the next stage involves decoding these normalized values into concrete RF parameters that can be realized within the physical environment. Each element in the matrix $\boldsymbol{D}$ represents a directional component that has been optimized in a normalized form. In order to translate these values into actionable wavefront directions, a decoding mechanism is required to convert the normalized entries into physical azimuth and elevation angles suitable for control by the PWE. This translation step is critical, as it enables the abstract solution derived from the optimization phase to be instantiated as a physical RF wavefront through the coordinated behavior of the spatially distributed SDMs. To construct this decoding process in a principled way, it is first necessary to identify the feasible angle-of-arrival range for each receiving antenna. This range is determined by the physical placement of the antennas within the environment, as well as the spatial distribution and steering capabilities of the SDMs. More concretely, every antenna is associated with a spatial support region that defines the bounds within which it can effectively receive incoming signals. These bounds are characterized by two angular intervals per antenna, one for azimuth and one for elevation, which together form what is referred to as the antenna-PWE angle range. This range represents the operational angular space available for directing incident wavefronts at each antenna, and it is this space over which the normalized values in $\boldsymbol{D}$ are mapped. Through this mapping, the statistical structure encoded in the optimized matrix is converted into a set Direction-of-Arrival vectors.

However, it is important to recognize that not all angles within the antenna-PWE angle range are equally feasible in practice. Due to the physical characteristics of the environment and the layout of the SDMs, some angular spans are more readily achievable, i.e., they offer a higher likelihood of effective wavefront steering. This non-uniformity in angular accessibility introduces a practical consideration for the decoding process, namely that the mapping from normalized values should ideally prioritize those angular directions where steering is more robust and reliable. In that sense, to perform this mapping in a way that favors angular directions with higher feasibility, the angular coverage of each antenna is first represented through a density function that describes how easily each angle can be achieved under the current SDM deployment. This density function shows how easily the SDMs can steer waves in different directions, with higher values indicating regions where the SDMs can more reliably direct the wavefront. Based on this density, a cumulative function is constructed, and the normalized values from $\boldsymbol{D}$ are then mapped into real angles through the inverse of this cumulative function. This approach ensures that more resolution in the normalized domain is assigned to angular regions that are more accessible in practice, improving the physical realizability of the decoding process. Formally, the azimuth and elevation angles are decoded for each antenna as \begin{equation}
\phi = f(d_{i}^{j}) = F_{\phi}^{-1}(d_{i}^{j}),
\end{equation}
\begin{equation}
\theta = g(d_{i}^{k}) = F_{\theta}^{-1}(d_{i}^{k}),
\end{equation}
where $d_{i}^{j}$ and $d_{i}^{k}$ are the normalized DoA entries for azimuth and elevation respectively, while $F_{\phi}^{-1}(\cdot)$ and $F_{\theta}^{-1}(\cdot)$ denote the inverse cumulative density functions of the azimuth and elevation angle ranges. Once decoded, the directional structure of the RF-Reading is recovered through the vectors \begin{equation}
\tilde{\boldsymbol{w}}_{\phi} = [\phi_1,..., \phi_m],
\end{equation}
\begin{equation}
\tilde{\boldsymbol{w}}_{\theta} = [\theta_1, ..., \theta_m],
\end{equation}
where $m$ is the total number of receiver antennas, and $\phi_i$ and $\theta_i$ represent the actual target azimuth and elevation angle corresponding to the $i$-th antenna, respectively. This decoding process completes the link between the abstract solution of the optimization and the physical RF wavefront to be synthesized, enabling the generation of directional RF-Readings that are consistent with both the statistical target and the operational constraints of the environment.

\begin{algorithm}[t!]
    \caption{The DoA-Matrix Encoding Algorithm}
    \label{alg:rf_vector_opt}
    \begin{algorithmic}[1] 
        \State \textbf{Input:} The target correlation matrix $\bm{R}$; the desired number of iterations, $\textit{wantedIterations}$.
        
        \Statex 
        \State /* \textbf{Initialization.} */
        \State Initialize mean vector $\bm{\mu} \gets \bm{0}$
        \State Initialize covariance matrix $\bm{\Sigma} \gets \bm{R}$
        \State Sample the initial matrix $\bm{D_0} \sim \mathcal{N}(\bm{\mu}, \bm{\Sigma})$
        
        \Statex
        \State /* \textbf{Iterative Optimization.} */
        \While{$i < \textit{wantedIterations}$}
            \For{each row $\bm{d}$ in $\bm{D}$}
                \State Update the $\bm{d}$ entry using L-BFGS-B
            \EndFor
        \EndWhile
        \State \textbf{Output:} $\bm{D}$
    \end{algorithmic}
\end{algorithm}

\section{PWE Routing and Visual Reconstruction}

Building upon the RF-Encoded DoA-vectors, the subsequent step involves translating them into routing instructions within the PWE. This process utilizes the deployed SDMs to orchestrate the propagation of RF signals, thus realizing the suitable RF-Readings at the inspection interface. In this direction, this section details the methodology for determining the appropriate routing algorithm to achieve the desired directional propagation, and finally mapping the encoded RF-Readings into visual inspection data through a GAN model.

\subsection{PWE-Routing}

\begin{figure}
\centering
\begin{subfigure}{0.95\linewidth}
\includegraphics[width=\textwidth]{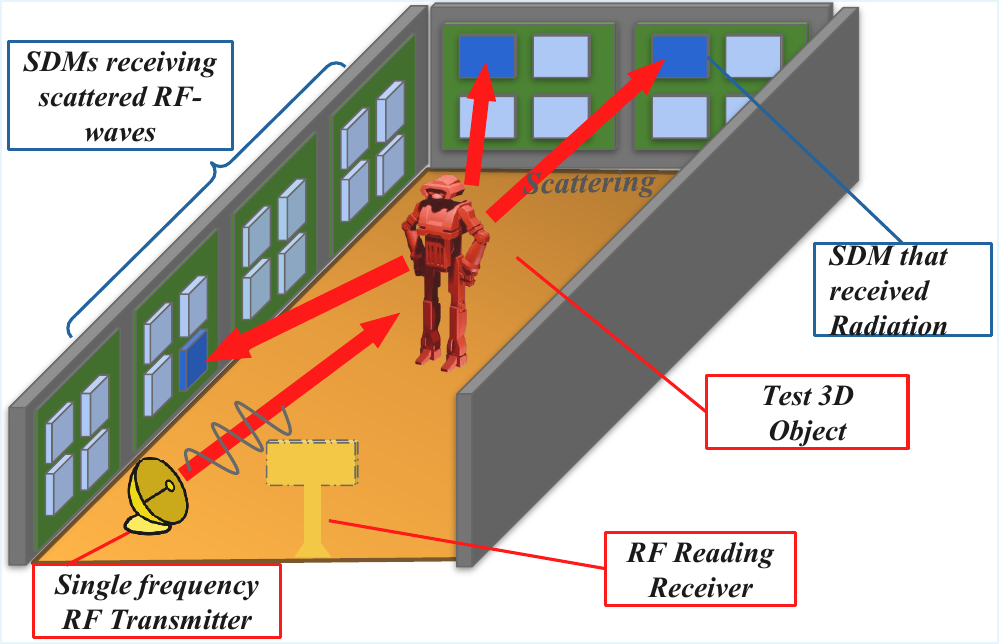}
\end{subfigure}

    \vspace{0.3cm} 
    (a) Ray-Tracing Scattering Behavior
    \vspace{0.3cm} 

\begin{subfigure}{0.95\linewidth}
\includegraphics[width=\textwidth]{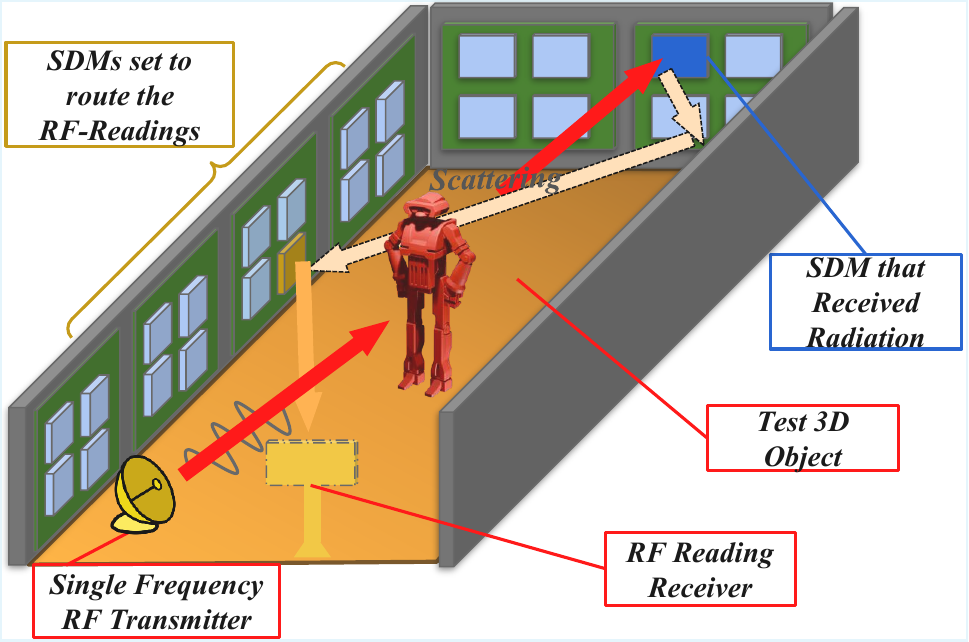}
\end{subfigure}

    \vspace{0.3cm} 
    (b) PWE-Routing 
    \vspace{0.05cm} 

\caption{Examples of Ray-Tracing and Ray-Routing in the PWE environment.}
\label{fig: Ray-Traceing-Routing}
\end{figure}

To enable the targeted delivery of RF signals that conform to desired directional characteristics, the PWE must support flexible and tractable control over wave propagation paths. A useful perspective for achieving this control is to model the PWE as a graph-based structure $\mathcal{G} = (\mathcal{V}, \mathcal{E})$, where the underlying connectivity and routing behavior of EM waves can be abstracted in algorithmically manageable terms. In this formulation, the set of graph vertices, $\mathcal{V}$, consists of the reconfigurable SDM elements along with the fixed transmitter and receiver locations, while the edges, $\mathcal{E}$, represent LoS propagation links between physically visible nodes, as described in \cite{liaskos2019network}. Crucially, the creation and connectivity of this graph is inherently tied to the SDM codebooks, since each SDM’s ability to establish a link, by steering an incoming signal, is constrained  by the available functionalities dictated by the codebook according to
\begin{equation}
    f_\text{codebook}(v_i, v_j) \in \{0,1\}, 
\end{equation}
where $f_{codebook}(v_i,v_j)$ denotes a feasibility function that evaluates whether a wave can be steered from node $v_i$ to node $v_j$ using the available SDM functionalities. Under this representation, the routing of signals through the PWE can be treated as a path-finding problem over the graph, where the objective is to direct the signal from the transmitter to a set of designated receiver antennas such that specific angle-of-arrival criteria are satisfied. These criteria are defined by the decoded DoA vectors obtained from the directional encoding process and must be respected in the final wavefront configuration. As a result, the PWE routing task becomes a constrained path-planning procedure, where the structure of the environment is leveraged to enforce signal delivery consistent with the learned RF-Reading characteristics.

The first step in the RF-routing procedure takes place during the training phase, where the goal is to determine the initial SDMs that receive RF energy scattered off the industrial asset. This is feasible because the assets involved in this phase originate from a known dataset, with fixed geometries and predefined sets of random orientations. Given this setup, a ray-tracing framework is used to simulate how RF signals, emitted by the transmitter, interact with the surface of the asset depending on its pose and spatial position. These simulations provide detailed insight into the scattering behavior, revealing which SDMs first receive the reflected wavefronts. Once identified, these SDMs are designated as the starting nodes of the routing process, as they represent the spatial entry points of the scattered RF energy into the PWE. An example of this interaction, including the propagation and the corresponding signal impact regions, is illustrated in Fig. 2a, highlighting how the scattering defines the relevant SDM subset for subsequent directional control.

Upon establishing the set of initial SDMs, the next step involves routing the impinging wavefronts toward the receiver antennas in a way that ensures compliance with the decoded directional targets. This is enabled through the graph-based abstraction of the environment, where SDMs, transmitter, and receiver elements are modeled as nodes, and LoS connections define the graph edges. Within this structure, the previously identified entry set of SDMs serve as the graph’s sources, while the receiver antennas act as destination nodes. The routing objective is to guide the wavefront along feasible paths so that it arrives at the receiver with a direction-of-arrival profile consistent with the decoded values. This is achieved by coordinating the behavior of intermediate SDMs and determining the final SDM that performs the last directional steering operation toward the receiver. The routing process is dictated by the algorithm introduced in \cite{ieeeBalkan2024}, which accounts for spatial constraints, steering feasibility, and angular precision. The overall mechanism is demonstrated in the Fig.2b, where the path taken by the wavefront and the involved SDMs are depicted as they contribute to constructing the target RF-Reading.

\subsection{Visual Inspection Reconstruction} \label{sec:GRAPH}

Following the completion of the RF-routing process, wherein wavefronts are guided through the PWE to reach the designated receiver antennas, the received directional information must be translated back into the normalized domain used during the optimization phase. This step ensures that the statistical structure initially optimized for preserving visual similarity, corresponding to structural topology, is recovered from the physical signal that has been shaped by the PWE. Specifically, the real azimuth and elevation angles received at the antennas, denoted by $\tilde{\boldsymbol{w}}_{\phi}$ and $\tilde{\boldsymbol{w}}_{\theta}$, are converted into normalized values within the $[0, 1]$ interval through the inverse decoding functions $f^{-1}(\cdot)$ and $g^{-1}(\cdot)$, yielding the vectors $\boldsymbol{\hat{\phi}}$ and $\boldsymbol{\hat{\theta}}$. These operations restore the directional components to the statistical domain in which the original correlation matrix was defined, thereby maintaining the consistency of the signal semantics across the system pipeline. Once re-normalized, the directional features are augmented with additional propagation-related parameters measured at each antenna, namely the path-loss $l_i$ and the phase-difference $\psi_i$, resulting in a complete RF-Reading of the form $\boldsymbol{\hat{w}} = [\hat{\phi}_1, \hat{\theta}_1, l_1, \psi_1, \ldots, \hat{\phi}_m, \hat{\theta}_m, l_m, \psi_m]$. This comprehensive representation captures both the structured spatial information and the environment-induced signal characteristics associated with each receiving element.

However, despite the controlled nature of the PWE and the correlation-preserving design of the DoA-vectors, practical limitations prevent the received RF-Readings from precisely matching their ideal counterparts. These discrepancies stem from quantization effects and discrete actuation constraints imposed by the SDM codebooks,  which restrict how accurately wavefronts can be shaped. As a result, the system cannot rely on a deterministic one-to-one mapping from RF-Readings to asset images. To address this inherent ambiguity, the translation from signal to visual reconstructions is approached through a generative modeling framework that learns to infer plausible asset reconstructions from imperfect but feature-rich signal inputs. In particular, a GAN is employed to capture the complex, data-driven relationship between directional RF features and visual structure. Specifically, the pix2pix conditional GAN framework introduced in \cite{pix2pix2017} is adopted for image-to-image translation tasks. This model consists of a U-Net generator and a PatchGAN discriminator and it has been shown to be effective for structured image synthesis tasks, which aligns with our task of generating shape-preserving asset reconstructions from structured signals. To enable compatibility with the GAN input format, the RF Reading $\hat{\boldsymbol{w}}$ is decomposed into four sub-vectors, corresponding to the azimuth, elevation, path-loss, and phase components, each of which is then mapped to a sub-image via a dedicated colormap transformation 

\begin{equation}
    \hat{\boldsymbol{w}}_c \xmapsto{\text{colormap}} \mathcal{I}_c(\hat{\boldsymbol{w}}_c),
\end{equation}
where $\hat{\boldsymbol{w}}_c$ is the decomposed sub-vector and $\mathcal{I}_c(\hat{\boldsymbol{w}}_c)$ is its corresponding sub-image representation. These sub-images are vertically stacked to construct a composite RF-based image $\mathcal{I}(\hat{\boldsymbol{w}})$ that serves as the GAN input. This representation preserves the spatial consistency of the signal modalities and provides a structured encoding format suitable for visual synthesis. During training, each RF-based input image is paired with its corresponding ground-truth asset image to form a dataset of aligned tuples $[\mathcal{I}(\hat{\boldsymbol{w}}), \text{VisualTwin}]$, enabling the GAN to learn the latent mappings between directional signal features and object appearance. To this end, the system can generate visually coherent representations of industrial assets based solely on the structured signal features received through the PWE.

\section{Simulation Results}
\label{sec: results}

\begin{figure*}[htbp]
\centering
 \includegraphics[clip,width=0.9\linewidth]{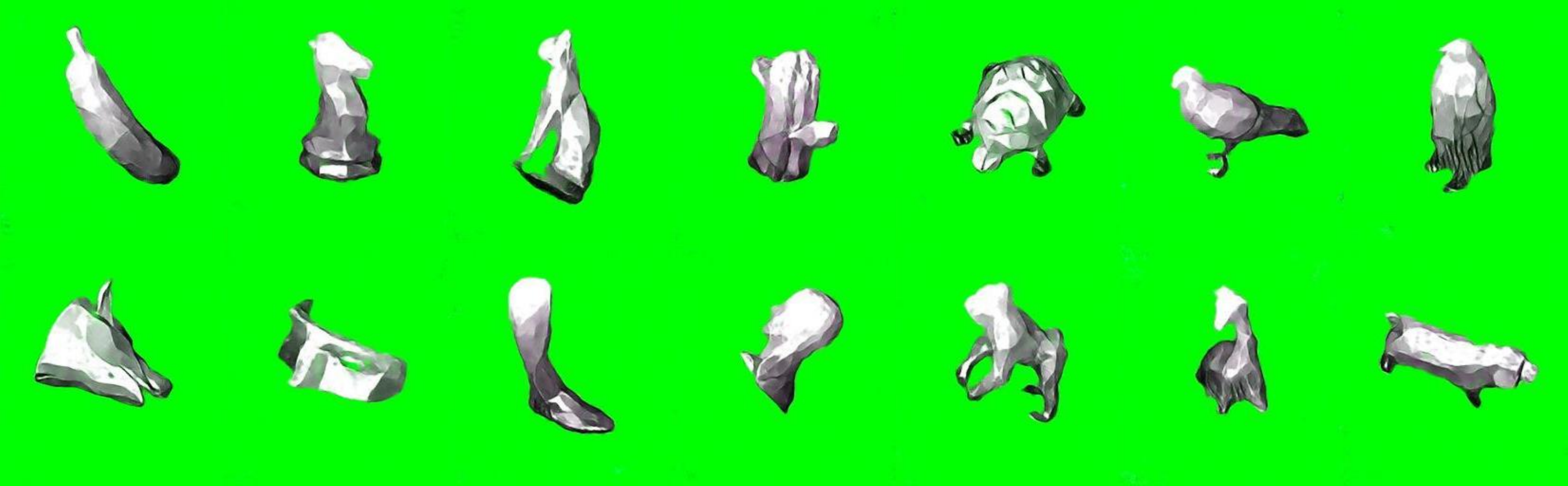}
\caption{\protect\label{fig:results} Indicative results of Digital Twin visuals generated by the pix2pix model.}
\end{figure*}

\begin{figure}[t!]
\centering
\begin{subfigure}{0.9\linewidth}\label{fig:morphing}
\includegraphics[width=\textwidth]{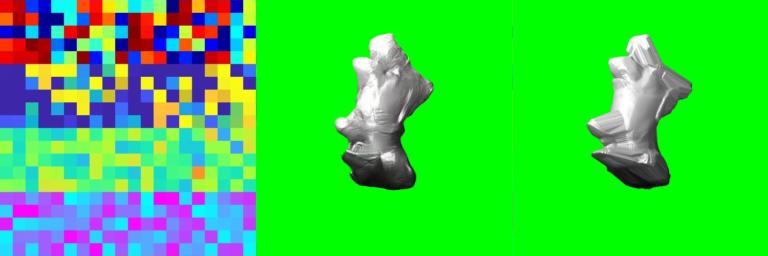}
\end{subfigure}

    \vspace{0.1cm} 
    (a) Morphing example
    \vspace{0.1cm} 

\begin{subfigure}{0.9\linewidth}
\includegraphics[width=\textwidth]{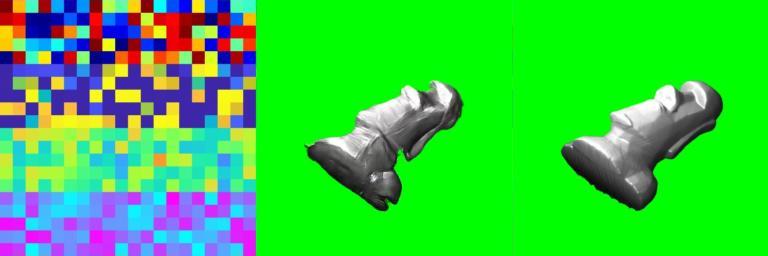}\label{fig:normal}
\end{subfigure}

    \vspace{0.1cm} 
    (b) Simple geometry example
    \vspace{0.1cm} 

\caption{Examples of GAN's results, where Leftmost: $\mathcal{I}(\hat{\boldsymbol{w}})$, Middle: GAN output, Rightmost: Ground truth.}
\label{fig: Gan Results}
\end{figure}

\begin{figure}[t!]
\centering
\begin{subfigure}{0.9\linewidth}
\includegraphics[width=\textwidth]{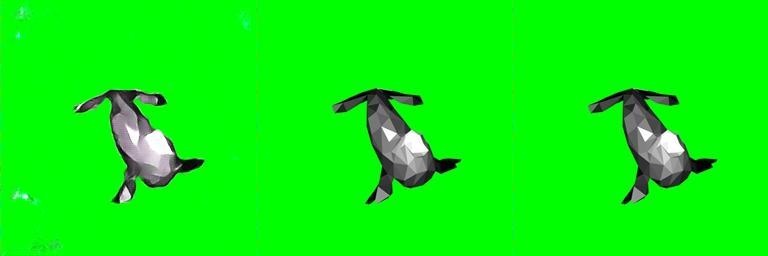} 
\end{subfigure}

    \vspace{0.05cm} 
    (a) Angle mismatch
    \vspace{0.1cm} 

\begin{subfigure}{0.9\linewidth}
\includegraphics[width=\textwidth]{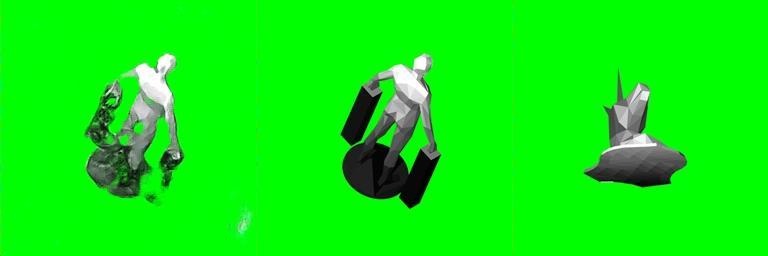}
\end{subfigure}

    \vspace{0.1cm} 
    (b) Object mismatch
    \vspace{0.05cm} 
    
\caption{Examples of GAN-metric mismatches, where Leftmost: GAN output image, Middel: Ground truth database image, Rightmost: Matched database image.}
\label{fig: Gan Mismatch}
\end{figure}

In this section, we evaluate the performance of the proposed PWE-based RF-to-image framework, by conducting a series of ray-tracing based simulations within a controlled indoor environment designed to emulate realistic signal propagation, i.e., multipath consideration and object induced EM interactions. The simulation scenario consists of an enclosed room with dimensions $(x=5, y=8, z=3)$ meters, where all walls are coated with programmable SDMs of size $d \times d$, with $d = 10$ cm. These SDMs enable dynamic manipulation of impinging RF signals according to the predefined codebook functionalities discussed. Additionally, four single-frequency RF emitters, operating at $f = 5$ GHz, are arranged in a rectangular layout at the following coordinates: $(-2, -3.5, 1.75)$, $(-2, 3.5, 1.75)$, $(2, -3.5, 1.75)$, and $(2, 3.5, 1.75)$, while a $10\times10$ array of receiver antennas is positioned at $(0, -4, 1.5)$ meters with $\lambda/2$ spacing. Especially,the receiver's set up is a dual-polarization one, hence considering a holistic RF information reception. Furthermore, the scene includes a 3D object selected from a predefined database, positioned at the center of the room. The objects are modelled as STL 3D geometries, which are compatible with the MATLAB ray-tracing tool, used to simulate the propagation and scattering of the RF-signals. Furthermore, considering the PWE-routing step, MATLAB-based graph solvers are used to compute feasible signal paths through the PWE graph. Subsequent to the routing process, the steps dictated in Subsection \ref{sec:GRAPH} are followed to achieve the translation of the RF-Reading to Digital Twin reconstructions. Specifically, the conversion of the RF-Readings' components $\hat{\boldsymbol{w}}_c$ to the sub-picture representations $\mathcal{I}_c(\hat{\boldsymbol{w}}_c)$ adhere to the 3000 resolution colormap encodings, of "jet", "parula", "turbo", "cool" for the azimuth, elevation, path-loss and phase-difference components, respectively. Finally, the training of the "pix2pix" model is performed for 200 epochs with a batch size of 64, using the Adam optimizer with learning rate $lr=0.0002$ and $\beta_1 = 0.5$. Overall, the final dataset utilized to train the "pix2pix" model is comprised of 40 different objects, illustrating various arbitrary geometries of surrogate industrial entities. These are, for example, various products, free-form components, complex castings etc., where each one of them is rotated 100 times over random various axes and angles of rotation, corresponding to a total of 4000 dataset samples. Furthermore, a second, differently derived, dataset was chosen to be studied, in which 12 different objects are morphed into one another over 100 morphing steps, resulting in an additional 6612 dataset entries, which are also randomly rotated. This very choice was made to simulate manufacturing variations and geometric deviations. In particular, 12 real-life objects were chosen to be the reference shapes, and based on a morphing percentage the shape of the more complex geometry is gradually transformed accordingly to an intermediate geometry closer to the simpler one. These new arbitrary shapes enrich the dataset’s capacity to incorporate deformed geometries, thus enhancing the model's generalization capability to learn patterns that link RF-Readings to variations-sensitive visual reconstructions. Some indicative results of the PWE-based RF-to-image framework can be seen in Fig. \ref{fig:results}. Although the target and generated images are grayscale, they are rendered against a green background to improve visual clarity in the displayed figures.

In Fig. \ref{fig: Gan Results}, examples of the GAN's inputs and outputs are illustrated. In particular, in the leftmost image the input image is depicted, while in the rightmost image the ground truth image is displayed, and finally at the center lies the output of the GAN model. Specifically, regarding the case of Fig. 3a an arbitrary morphing is depicted, while in Fig. 3b a simple 3D geometry, of a random object, is illustrated. Based on the outputs of the "pix2pix" it is possible to provide the initial database Digital Twin reconstruction to the inspector, by implementing a similarity test among the generated GAN images and the database ones. The similarity task takes place by comparing the GAN generated images with the database target images through the utilization of various image similarity measures, e.g. Euclidean L2 norm, mutual information \cite{russakoff2004image}, cosine similarity, peak signal-to-noise ratio (PSNR) \cite{sara2019image} and structural similarity index measure (SSIM) \cite{wang2004image}. By comparing all those measures it can be deduced, as it is evident on the Table \ref{tab:my_scores}, that the SSIM approach performs better than the other ones in providing the correct database DT data, with regard to the GAN outputs, with an accuracy score of $99.5\%$. 

In Fig. \ref{fig: Gan Mismatch} two mismatching phenomena are depicted, highlighting the case that the similarity task might fail, at providing the correct database entry to the inspector. Specifically, in the aforementioned figure the GAN output is positioned on the left hand side and the matched, by the similarity metric, database image is depicted on the right hand side, while the ground truth database image is illustrated on the middle. In more detail, a distinction regarding the mismatches that are highlighted by the similarity metrics can be made. As it is evident in the Fig. 5a and Fig. 5b there are two categories of mismatches that the measures of similarity can designate considering the model's performance. The first one is an angle mismatch, where the GAN model generates an output image that is really close to the target one but there exists an entry in the database, of the same object, that is even closer to the GAN output. That is an aftereffect of the databases's data, which  sometimes are differentiated by small 3D rotational differences of the object. That results in a false association of the RF-Reading to a non-appropriate target, but nevertheless the output and the target are still really close images, preserving the correct identification of the asset type. On the other hand, the model might not sufficiently manage to generate the correct target image. Thus the similarity measures that are used to match the GAN model's outputs to database's entries will result to matching images of wrong structural topology. This scenario also leads in a false association of an RF-Reading to a wrong target, yet this time the output and the target will be very different images. It can be realized that the first category of mismatches is not of a great degrading factor to our goal, but in the second case if a lot of object mismatches are observed then the objective of automated NDI is undermined. Because of that the main concern in the usage of a similarity measure is to be also able to erase any false association of the "object mismatch" category. Thus, an other important parameter to be considered is how well the similarity measures accomplish to minimize the object mismatches. This is shown in Table 2, where higher scores of angle mismatches mean better performance, while ideally no object mismatches should be present. Thus, aggregating the results observed at Table 1 and Table 2 we can deduce that the SSIM metric is the best performing one, as it manages to minimize the object mismatch score, while at the same time maximizing the matching score between the GAN outputs and the database entries.

   \begin{table}[ht]
    \centering
    \resizebox{0.75\linewidth}{!}{
    \begin{tabular}{c|c}
        \hline
        \textbf{Metric} & \textbf{Matching Score} \\ \hline
        SSIM & 99.5\% \\ \hline
        L2 Norm & 99.38\% \\ \hline
        PSNR & 99.38\%  \\ \hline
        Mutual Information & 99.32\% \\ \hline
        Cosine Similarity & 99.32\%
    \end{tabular}
     }
    \caption{Matching scores of similarity metrics at providing the correct database Digital Twin visuals to the inspector.}
    \label{tab:my_scores}
\end{table}

\begin{table}[ht]
    \centering  
    \renewcommand{\arraystretch}{1.2}
    \resizebox{1\linewidth}{!}{
    \begin{tabular}{c|c|c}
        \hline
        \textbf{Metric} & \textbf{Angle Mismatch} & \textbf{Object Mismatch}\\ \hline
        SSIM & 100\% & 0\% \\ \hline
        L2 Norm & 95.7\% & 4.3\% \\ \hline
        PSNR& 95.7\% & 4.3\%  \\ \hline
        Mutual Information & 100\% & 0\% \\ \hline
        Cosine Similarity & 96\% & 4\%
    \end{tabular}
    }
    \caption{Mismatching scores of similarity metrics at providing database entries that classify as angle mismatch, or asset mismatch.}
    \label{tab:my_mismatches}
\end{table}

\section{Conclusion}

In this work, a novel application of the PWEs was studied, presenting how exerting control over the RF waves' propagation can be utilized for generating visual outputs tailored for Non-Destructive Inspection. Through the synergetic operation of the Software Defined Metasurfaces and a Generative Adversarial Network, a system that maps RF-wavefronts to Digital Twin reconstructions, with a matching score of $99.5\%$, was suggested. This study highlights the PWEs' versatile nature as it serves a role for enabling advanced, privacy-aware industrial monitoring pipelines. Future work could explore the idea of complete material-aware volumetric composition monitoring within a PWE for advanced Non-Destructive Inspection systems.

\ifCLASSOPTIONcompsoc
  \section*{Acknowledgments}
\else
  \section*{Acknowledgment}
\fi
This work has been funded by the FORTH Synergy Grant 2022 `WISAR' and the European Union's Horizon 2020 research and innovation programmes, CLIMOS (GA EU101057690), CYBERSECDOME (GA EU101120779).

\bibliographystyle{IEEEtran}
\bibliography{bibliography}

\end{document}